%% file: ICVGIP-Latex-Template.tex
\documentclass[sigconf]{acmart}

\usepackage{hyperref}
\usepackage{booktabs} 
\usepackage{soul}
\usepackage{multirow}
\usepackage{amsmath} 
\usepackage[]{algorithm2e}

\setcopyright{acmcopyright}

\acmConference[ICVGIP 2018]{11th Indian Conference on Computer Vision, Graphics and Image Processing}{December 18--22, 2018}{Hyderabad, India}
\acmBooktitle{11th Indian Conference on Computer Vision, Graphics and Image Processing (ICVGIP 2018), December 18--22, 2018, Hyderabad, India}
\acmYear{2018}
\copyrightyear{2018}
\acmDOI{10.1145/3293353.3293434}
\acmISBN{978-1-4503-6615-1/18/12}
\acmPrice{15.00}
\acmArticle{81}
\editor{Anoop M. Namboodiri}
\editor{Vineeth Balasubramanian}
\editor{Amit Roy-Chowdhury}
\editor{Guido Gerig}

\begin{document}
\title{Integrating Objects into Monocular SLAM: \\Line Based Category Specific Models}

\author{Nayan Joshi\footnotemark[1]}
\orcid{1234-5678-9012}
\affiliation{%
  \institution{IIIT Hyderabad, India}
}

\author{Yogesh Sharma\footnotemark[1]}
\affiliation{%
  \institution{IIIT Hyderabad, India}
}

\author{Parv Parkhiya}
\affiliation{%
  \institution{IIIT Hyderabad, India}
}

\author{Rishabh Khawad}
\affiliation{%
  \institution{IIIT Hyderabad, India}
}

\author{K Madhava Krishna}
\affiliation{%
  \institution{IIIT Hyderabad, India}
}

\author{ Brojeshwar Bhowmick}
\affiliation{%
  \institution{Tata Consultancy Services, India}
}


\renewcommand{\shortauthors}{K Madhava Krishna et al.}

\begin{abstract}
We propose a novel \emph{Line based} parameterization for category specific CAD models. The proposed parameterization associates 3D \emph{category}-specific CAD model and object under consideration using a dictionary based RANSAC method that uses object Viewpoints as prior and edges detected in the respective intensity image of the scene. The association problem is posed as a classical Geometry problem rather than being dataset driven, thus saving the time and labour that one invests in annotating dataset to train Keypoint Network\cite{newell2016stacked,parkhiya2018constructing} for different category objects. Besides eliminating the need of dataset preparation, the approach also speeds up the entire process as this method processes the image only once for all objects, thus eliminating the need of invoking the network for every object in an image across all images. A 3D-2D edge association module followed by a resection algorithm for lines is used to recover object poses. The formulation optimizes for shape and pose of the object, thus aiding in recovering object 3D structure more accurately. Finally, a Factor Graph formulation is used to combine object poses with camera odometry to formulate a SLAM problem.
\footnotetext[1]{Authors contributed equally.}

\end{abstract}

%
%


\maketitle

\input{samplebody-conf}

\bibliographystyle{ieeetr}
\bibliography{sample-bibliography}

\end{document}

%% file: samplebody-conf.tex
..
\section{Introduction}
\begin{figure}[!htb] 
 \centering
  \includegraphics[width=1\linewidth, height=5cm]{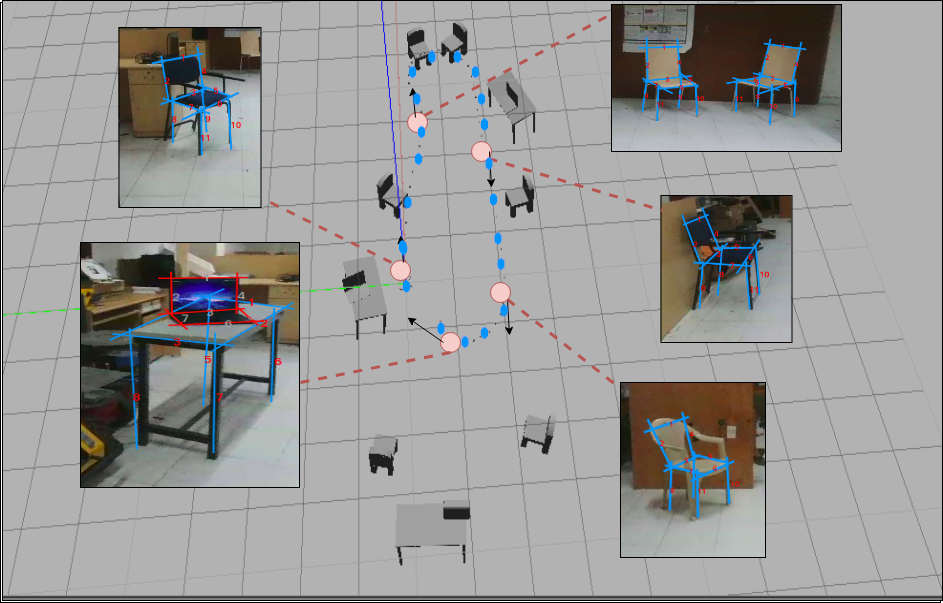}
 
  \caption{Qualitative visualization of our \emph{Line Based} Object-SLAM implemented on real monocular sequence. For a given RGB image taken from a monocular camera (read MAV), we estimate camera trajectory along with shapes and pose of different objects. }
   \label{Fig:SLAM_visual}
\end{figure}
Simultaneous Localization and Mapping (SLAM) is the most vital cog in various mobile robotic applications involving ground robots\cite{lategahn2011visual}, aerial\cite{ccelik2013monocular} and under water vehicles\cite{cho2017visibility}. Monocular SLAM has emerged as a popular choice given its light weight and easy portability, especially in restrictive payload systems such as \emph{micro aerial vehicles}(MAV) and hand held camera platforms.

SLAM has evolved in various flavors such as active SLAM \cite{leung2006active}, wherein planning is interleaved with SLAM, dynamic SLAM\cite{kundu2011realtime} which reconstructs moving objects and robust SLAM\cite{agarwal2015others}. Object SLAM\cite{SLAM++} is a relatively new paradigm wherein SLAM information is augmented with objects in the form of its poses to achieve more semantically meaningful maps with the eventful objective of improving the accuracy of SLAM systems.

Object SLAM presents itself in two popular threads. In first, instance specific models are assumed to be known apriori \cite{Choudhary} . In the second, a general model for an object is used such as ellipsoids and cuboids \cite{yang2018cubeslam} and \cite{SLAM++}. Relying on instance level models for various objects in the scene makes the first theme difficult to scale to various objects in the scene whereas general models such as cuboids do not provide meaningful information at the level of object parts and limit its relevance in application that require grasping and handling objects.

To overcome such limitations, \cite{parkhiya2018constructing} positioned their research as one that combines the benefits of both. In particular, \emph{category specific} models were developed in lieu of \emph{instance level} models, which retained the semantic potential of the former along with the generic nature of the later at the level of object category. However, reliance of \cite{parkhiya2018constructing} on a keypoint trained network for a particular category limits its expressive power as every new object category entails the estimation of a new network model for that category along with the concomitant issues of annotation, GPU requirement and dataset preparation. More specifically, in a scene that contains three object categories \cite{parkhiya2018constructing} is entailed to invoke three separate network models corresponding to each category to solve for the pose and shape of the respective category of the object.

Motivated by the fact that many objects can be represented as line structures, this paper presents a novel line parameterization of objects for an object category. By associating 3D line that characterize the object category in 3D and its observation in the image in the form of 2D line segment, we solve for the object pose and shape in a decoupled formulation.

Significantly, this approach bypasses the need for keypoint annotation as we expand our pipeline to new categories as well as the requirement of estimating and maintaining an assortment of network models for various category of objects. It achieves this by relying on line segment detectors for observation of object line segments in the image rather than network models trained for semantic keypoints.

The paper shows the scalability of the line parameterized objects to three categories (chair, table and laptop) and successfully integrates the shape and pose optimized object with a factor graph based backend pose-graph optimization. Thereby, it successfully embeds 3D objects into the scene while simultaneously estimating the camera trajectory. High fidelity estimation of camera trajectory and object poses vindicates the efficacy as well as the novelty of the proposed framework.

Fig \ref{Fig:SLAM_visual} Shows a typical Object SLAM run with the object poses rendered in 3D as the closest CAD model corresponding the optimized wireframe meshes shown in the inset image. Sample camera locations from the trajectory are shown in pink circles with the camera trajectory itself shown in the black dotted lines.

\section{Related Work}
Mostly, all state-of-the-art SLAM systems \cite{ORB,LSD,SVO, Edge} and reconstruction methods using IMUs \cite{sm1,sm2} rely on the pose-graph/ factor-graph optimization \cite{g2o,GTSAM} or bundle adjustment. In the following section we will review the related work on object-SLAM and discuss some limitations in them and the keypoint based approach which motivated for the proposed approach.
There are some approaches which tried to fuse the properties of classical geometry with deep learning models to improve object pose and shape. Latest in the line of such implementations is \cite{zhu2018object} which recover both global camera pose and 3D point cloud based shape with very few, limited view observations.  
\subsection{Object-SLAM}
Recent developments and the following stabilization of the SLAM systems, has led the community to incorporate objects into the SLAM framework and solve for object poses and shapes along with the robot poses in an unified framework. Some of the recent approaches for object-oriented SLAM are \cite{SLAM++,Paull,RAS2016}.

Majority of the object-based SLAM rely on depth information from RGB-D or stereo sensors. In \cite{crocco2016structure,Choudhary} instance level models are assumed, which is known as shape priori. In \cite{Choudhary}, a framework for multi-robot object-SLAM is proposed but again with a shape priori and RGB-D sensors.
In one other paradigm there is no instance-level models, available as priori. In \cite{Paull}, again with the help of RGB-D cameras, the association and object poses are solved jointly, in a factor graph framework. Among monocular objectSLAM/SfM approaches, \cite{crocco2016structure,RAS2016} fall under this paradigm. In such approaches, objects are modeled as bounding boxes \cite{RAS2016,Sunderhauf2015} or as ellipsoids \cite{crocco2016structure}.

Our proposed approach hence falls under a third paradigm, where we assume line based category-models, and not instance-level models. 

\subsection{Object-Category Models}
Over the last few years researchers have gradually started to re-introduce more and more geometric structure in object class models and improve their performance \cite{felzenszwalb2010object} . \emph{Object-category} model based approach is employed to solve various problems in monocular vision, in fact \cite{murthy2017reconstructing} - \cite{tulsiani2017learning} employed category-level models to reconstruct objects from single image. \cite{yang2018cubeslam} Propose a method for 3D cuboid object detection and multi-view object-SLAM without prior object models. They propose an efficient and accurate 3D cuboid fitting approach on single image, without prior knowledge of object model or orientation.  

\begin{figure}[!htb] 
 \centering
  \includegraphics[width=1\linewidth, height=2.5cm]{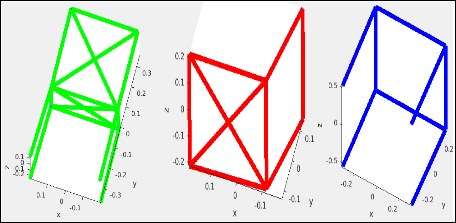}
 
  \caption{Our shape priors for chair, laptop and table. On varying the shape parameter($\wedge$) we can
deform the model to fit object shape.}
   \label{Fig:shape_prior}
\end{figure}

Approaches based on \emph{category-level} model advocate incorporating category specific shape priors of an object to compensate for information loss when dealing with monocular image based processing. We employ these models Fig.  \ref{Fig:shape_prior} to incorporate object observation factors into monocular SLAM by representing all instances of a category by same model,
\subsection{Object Detection and View Point Estimation}
Convolutional Neural Networks (CNNs) have been the driving factor behind the recent advances in object detection\cite{redmon2016you,ren2015faster,liu2016ssd}.
These CNNs are not only highly accurate, but are very fast as well. In fact when run on a GPU, they can process at a latency of 100-300 milliseconds for each image frame. Estimating good bounding boxes for object belonging to a specific category marks the outset of our architecture. \\
One such CNN based model is Render For CNN \cite{su2015render}, our proposed solution uses the same to estimate viewpoint of an object in an image. Render For CNN has been trained on large, category specific datasets for several objects, rendered using available 3D CAD models \cite{chang2015shapenet} that are easily accessible. Models that are trained for the task of object viewpoint prediction on rendered dataset work very well when they are fine-tuned on large dataset comprising of real images \cite{everingham2010pascal}.
\section{Methodology}\label{section:methodology}
In this section we explain the end to end functioning of our \emph{Line based} pipeline, giving detailed insight into each of the constituent stages. 

\subsection{Pipeline Overview}
\begin{figure*}[!htb]
 \centering
  \includegraphics[width=0.88\linewidth, height=10.5cm]{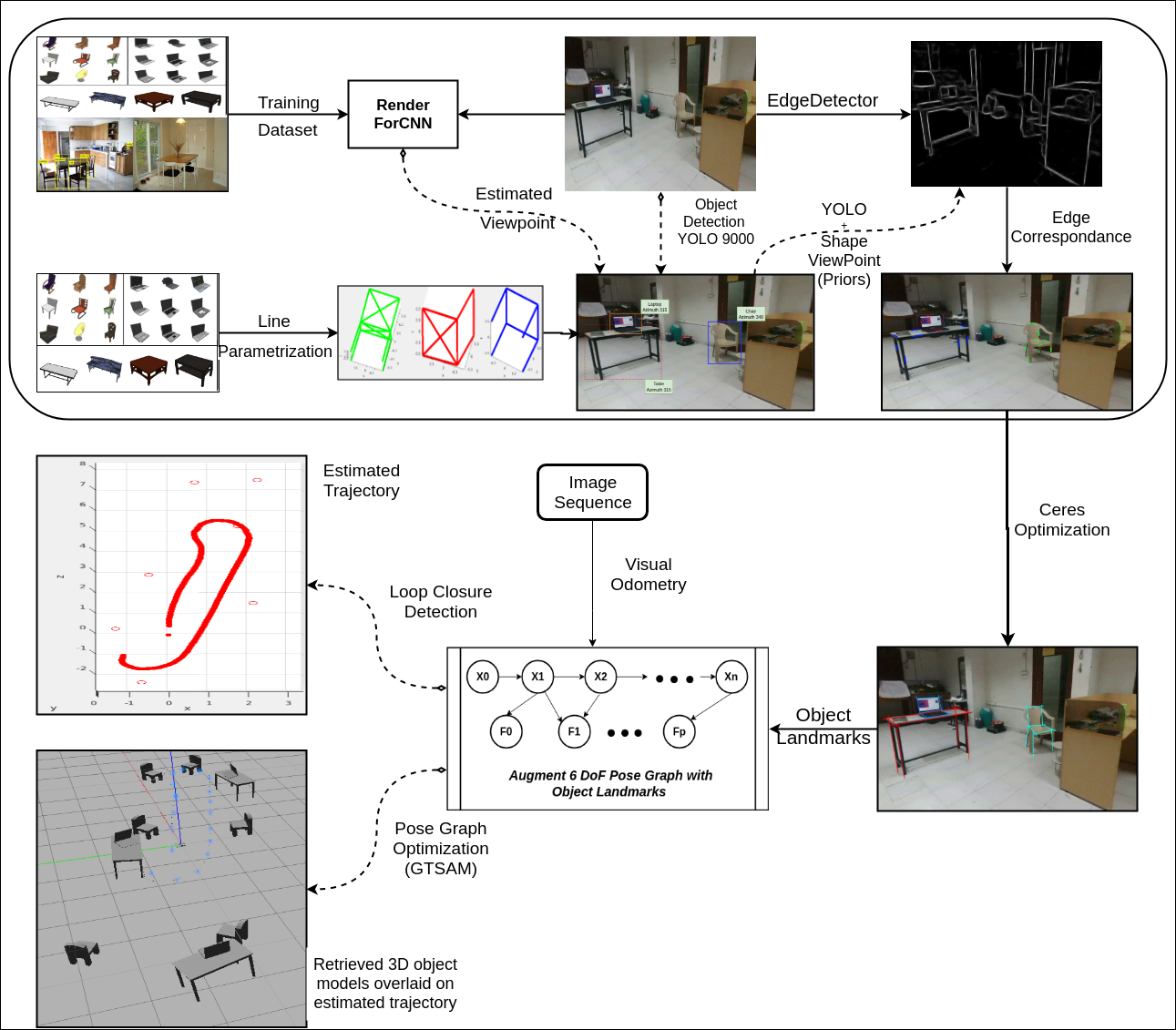}
  \caption{End to end pipeline of our architecture, incorporating learned  \emph{Category-Based} models for Object-SLAM.}
  \label{fig:complete_pipeline}
\end{figure*}

The render for CNN pipeline \cite{su2015render} is trained for category specific view point estimation of an object. When presented an image, YOLO detector \cite{redmon2016you} regresses bounding boxes on objects of interest. An LSD detector\cite{von2012lsd} outputs the line segments within the YOLO bounding boxes. The render for CNN model outputs the viewpoint prior. The data-association module associates lines of the mean wireframe model in 3D with the LSD observations of line segments within the bounding boxes. Subsequent to the data-association a pose-shape optimization module using Ceres Solver \cite{agarwal2015others} outputs the pose and shape of these objects. In a Object SLAM run the pose-shape optimization outputs constitute the camera pose-landmark constraint. Whereas the camera motion is estimated using state of the art SLAM module \cite{mur2015orb}. These constraints are finally optimized with GTSAM \cite{dellaert2012gtsam} as the backend engine to output the camera trajectory along with objects embedded in the scene. This pipeline is vividly portrayed in Fig \ref{fig:complete_pipeline} 

\begin{figure*}[!htb]
 \centering
  \includegraphics[width=0.95\linewidth, height=8.25cm]{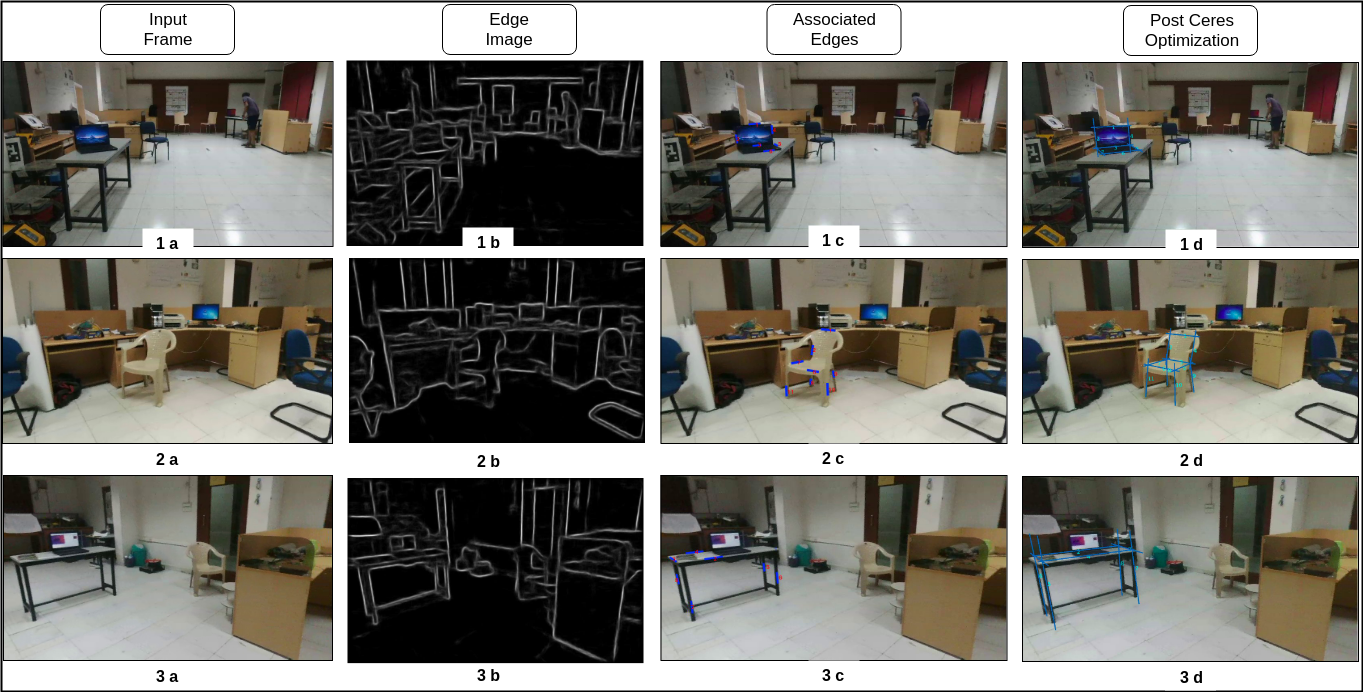}
  \caption{Qualitative visualization of our method, on 3 category of objects (laptop, chair and table). Column 1 represent scene as captured by robot (MAV). Column 2 represent intensity image of the scene that forms an input to our line based pipeline. Column 3 shows result of Edge Correspondence [\ref{sec:edge_correspondence}] on respective input frame. Once we have associated respective edges, Shape and Pose Optimization [\ref{sec:shape_pose_optimization}] yield accurate object shape and pose, as visualized in column no. 4. }
  \label{fig:lineBased_qualitative}
\end{figure*}
\subsection{Line based {\itshape Category-Level} Model} \label{subsection:line_parameterization}
In our approach, we lay an emphasis on the use of \emph{category-level} models as opposed to \emph{instance-level} models for objects. To construct a line based category level model, each object is first characterized as a set of 3D lines that are common across all instances of the category. For example, such lines for the \emph{chair} category could be legs of the chair, edges of the chair backrest, for laptops they can be the edges around the display screen and those contouring the keyboard, constituting the base and so on.   

Any line based model is represented by a vector $\mathbf{X}$ of 6*$m$ dimension, where $m$ is the number of lines present in the parameterized model, each $\mathbf{L_i}$ corresponding to a key edge of a model representing the object. Each of these $m$ lines is represented by a 3D direction $\mathbf{D}$ and a 3D point $\mathbf{M}$, one that lies on the line.
\begin{equation}
\mathbf{D} = <\mathbf{d}_x,\mathbf{d}_y, \mathbf{d}_z>
\end{equation}
\begin{equation}
\mathbf{M} = <\mathbf{m}_x, \mathbf{m}_y, \mathbf{m}_z>
\end{equation}
\begin{equation} \label{eq:line_param}
\mathbf{L}_i = <\mathbf{M}_i, \mathbf{D}_i>
\end{equation}
\begin{equation} \label{eq:mean_model_param}
\mathbf{X} = <\mathbf{L}_1, \mathbf{L}_2, \dots  ,\mathbf{L}_m>
\end{equation}
While the 3D point can be any point lying on the line, it is roughly chosen to be the midpoint of the edge of 3D CAD models. (e.g. midpoint of the leg of a chair)

If no prior information about object is known then search space is a prohibitive 6*$m$ dimensional space representing shape of the object. But based on the 3D annotation of CAD model, search space can be reduced so that while optimizing for shape only possible deformations in that object are looked at, rather than any arbitrary line deformation. A simple principle component analysis\cite{jolliffe2011principal} is performed on the annotated CAD model dataset to get the top seven linearly independent principle directions of the deformation. These eigen vectors are sorted based on their eigen values. The number seven is chosen based on the coverage of the eigen vectors. 

While solving for a shape, an object is represented by the mean shape plus \emph{weighted} linear combination of the deformation directions. In such a shape representation, each chair can be represented by those \emph{weights} (or \emph{shape parameters}, $\lambda_i$  ) for each principle deformation direction. This linear subspace model has much lower dimension than $\mathbb{R}^{6m}$. This is easy to see, since there are various planar conditions and symmetry present in the objects.   

Mathematically, if $\mathbf{\overline{X}}$ is the mean shape of the category, and $\mathbf{V_i}$s are a deformation basis obtained from PCA over a collection of aligned ordered 3D CAD models as explained in this section, any object $\mathbf{X}$ obtained with shape parameters $\lambda_i$ can be represented as, 
\begin{equation}
\mathbf{X} = \mathbf{\overline{X}} + \sum_{i=1}^{B} \lambda_i \mathbf{V_i} = \mathbf{\overline{X}} + \mathbf{V \Lambda}
\label{eqn:shapePrior}
\end{equation}
where $B$ is the number of basis vectors (the top-$B$ eigenvectors after PCA) and $\Lambda$ is vector consisting of all $\lambda_i$.

\subsection{Edge Correspondence} \label{sec:edge_correspondence}
Object invariant line detection is easier when compared to finding salient keypoints in non machine learning methods. We use LSD edge detector\cite{von2012lsd} to achieve the same. The main problem here arises in associating correct 2D lines, of all the lines detected, with the respective 3D lines.
Finding association is a chicken and egg problem in this case. We need a good pose estimation to find the correspondence between 3D CAD model and image and we need a good association to estimate pose of object.
We get an approximate viewpoint of the object using the RenderForCNN viewpoint\cite{su2015render} network and introduce a method to compute approximate translation of object. We employ this viewpoint and translation as initialization for a dictionary based RANSAC method to get most suitable Edge correspondences.

The parameterization discussed in section \ref{subsection:line_parameterization} allows for the representation of CAD models in terms of a set of vectors where each vector represents a line. To put it formally, we find correspondence map $Z$ from $n$ 3D lines to $m$ 2D line segments.
First, the line segments in image are filtered using the bounding box data we have from Object detector\cite{redmon2016you}.
We use a custom cost function to give a score to a 3D-2D correspondence
\begin{equation} \label{eq:edge_association_cost}
\mathbf{C = C_1 + k_1\times C_2 + k_2\times C_3}
\end{equation}
where, $C_1$ accounts for angle and $C_2$ and $C_3$ account for the distance between the line and line segment. In following subsections, we discuss the method to compute translation and the aforementioned costs.

\subsubsection{Computing Translation}
Apart from a viewpoint initialization, an approximate value of translation $(T_x, T_y, T_z)$ is also needed for projection. Getting exact translation requires 3D length and projected 2D length of a line segment, but since the exact 3D information of object is not known, we need to rely on approximation of 3D model of that particular category of object.

We use the information available from bounding box and mean 3D model to find translation approximation. Height and Width of bounding box are independently sufficient to get a good estimate of $T_z$ given that object's mean 3D model's height and width matches mean model respectively. In order to get even better estimate in general case where both height and width of objects could deviate from mean model, we simply take average of both estimate. 

\[T_z' = k_x\times\frac{f_x}{w}\]
\[T_z'' = k_y\times\frac{f_y}{h}\]
\begin{equation}\label{eq:computing_Tz}
 	\mathbf{T_z = \frac{T_z'+T_z''}{2}}
\end{equation}
\begin{equation} \label{eq:computing_Tx}
	\mathbf{T_x = (x + \frac{w}{2} - u)\times  \frac{T_z}{f_x}}
\end{equation}
\begin{equation} \label{eq:computing_Ty}
	 \mathbf{T_y = (y + \frac{h}{2} - v)\times \frac{T_z}{f_y}}
\end{equation}

Here, $f_x ,f_y, u, v$ are taken from camera matrix, $h, w$ are the height and width of bounding box and $x$ and $y$ are the top left corner of bounding box. $k_x$ and $k_y$ are constants obtained from mean 3D model. 

\subsubsection{computing $C_1$, $C_2$, $C_3$}
 From eq \ref{eq:line_param}
 \[L_i = <m_s, m_y, m_z, d_x, d_y, d_z>\] 
 \[or\ L_i = <\overline{M}, \overline{D}>\]

\begin{figure}[!htb]
\includegraphics[width=0.9\linewidth, height=5.5cm]{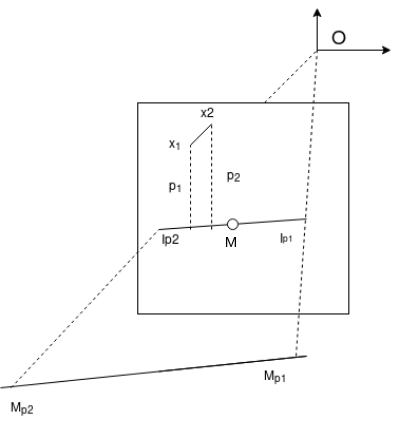}
 \caption{Projection of 3D line to 2D to calculate association cost with 2D line $x_1$ $x_2$}
 \label{Fig:edge_correpondence_image}
\end{figure}

The projection of the 3D edge to image plane can be found by projecting any two points from the 3D line and then taking their direction and mid point (See Fig \ref{Fig:edge_correpondence_image}). R and T are rotation and translation of the 3D line.
\[\overline{M_1} = \overline{M} + \alpha_1\times \frac{D}{2\times|D|}\]
\[\overline{M_2} = \overline{M} - \alpha_1\times \frac{D}{2\times|D|}\]
\[\overline{M_{p1}} = K\times(R\times\overline{M_1} + T)\]
\[\overline{M_{p2}} = K\times(R\times\overline{M_2} + T)\]
\[\overline{I_{p1}} = \pi(\overline{M_{p1}})\]
\[\overline{I_{p2}} = \pi(\overline{M_{p2}})\]
\[\overline{M} = \frac{\overline{I_{p1}} + \overline{I_{p2}}}{2}\]
here, $\alpha_1$ is some non-zero number used to get two points on line based on one point, $M$ and direction, $D$ and $\pi$ is the projection function.

    In fig. \ref{Fig:edge_correpondence_image}, $x_1$ and $x_2$ are the end points of an edge detected by LSD (for a line segment to be categorized as associated line, it has to be very close to the projected 3D line. The image here is exaggerated for representation purpose) and $I_{p1}$ and $I_{p2}$ are the projection of two points from 3D line. $p_1$ and $p_2$ are the perpendicular distances when $\overline{x_1 x_2}$ is projected on $\overline{I_{p1} I_{p2}}$. Using simple projective geometry, we get,

\begin{equation} \label{eq:perpendicular_1}
 \mathbf{p_1 = \frac{\overline{x_1} \odot (\overline{M_1}\otimes \overline{M_2})}{|\overline{M_1}\otimes \overline{M_2}|}}
\end{equation}

\begin{equation} \label{eq:perpendicular_2}
 \mathbf{p_2 = \frac{\overline{x_2} \odot (\overline{M_1}\otimes \overline{M_2})}{|\overline{M1}\otimes \overline{M_2}|}}
\end{equation}

Adding angle directly in cost function would create complication of adding distance with angles so instead we observe that value $|p2 - p1|$ captures the variation of angle between the two lines. This is used as $C_1$
\begin{equation}\label{eq:c1}
C_1 = |p_2 - p_1|
\end{equation}

$\frac{p1 + p2}{2}$ captures the perpendicular distance of the midpoint of $\overline{x_1 x_2}$ from the projected line $\overline{I_{p1} I_{p2}}$. This is used as $C_2$
\begin{equation} \label{eq:c2}
C_2 = \frac{p_1 + p_2}{2}
\end{equation}

and lastly, distance between $M_p$ and $\frac{x_1 + x_2}{2}$ is minimized to pick the lines radially closer to the projected line. This is used as $C_3$
\begin{equation} \label{eq:c3}
C_3 = euclidean\_distance(M, \frac{x_1 + x_2}{2})
\end{equation}

so using eq \ref{eq:c1}, \ref{eq:c2} and \ref{eq:c3}, the final association cost in equation \ref{eq:edge_association_cost} becomes
\begin{equation}\label{eq:total_association_cost}
\mathbf{C = |p_2 - p_1| + k_1\times\frac{p_1 + p_2}{2} + k_2\times{distance(M, \frac{x_1 + x_2}{2})}}
\end{equation}

\subsubsection{Association Pseudocode} We generate a dictionary of 3-5 most representative CAD models (selected manually) for each category of object, represented by $\overline{X_D}$. Also, we sample viewpoint around azimuth initialization and translation around the computed $T$ for $RANSAC$. Let's call the sampled set $\overline{V_p}$ and $\overline{T_s}$, respectively.

Now, we can write the pseudocode for our $RANSAC$ based association algorithm which iterates over dictionary models and sampled view points, projects them and calculate associated lines and cost of association. Association for a line in model with a view point and translation is the line segment in image which has the minimum cost corresponding to that line in model.

Finally, it picks the association pertaining to the lowest association cost. see, $Algorithm\ 1$
\begin{algorithm}
\SetAlgoLined
\For{$x$ in $\overline{X_D}$} {
	\For{$v_p$ in $\overline{V_p}$} {
    	\For{$t$ in $\overline{T_s}$} {
        	$Project(x, v_p, t)$\;
            Find association cost, $c_i$ and associated lines $a_i$\;
            $Costs.append(v_p, t, c_i)$\;
        }
    }
}
[$Associated\_lines$, $cost$] = $min_c(Costs);$
\caption{Association Pseudocode}
\end{algorithm}

\subsection{Pose and Shape Optimization}
\label{sec:shape_pose_optimization}
\begin{figure}[!htb]
 \centering
  \includegraphics[width=0.8\linewidth, height=5cm]{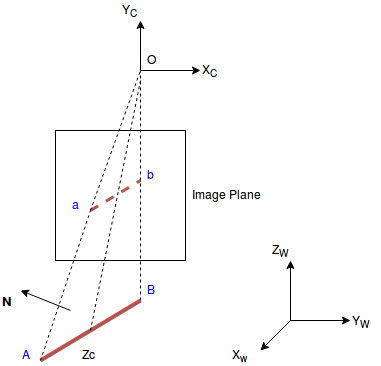}
  \caption{Perspective Projection of Image Lines}
  \label{Fig:ceres_line_resection}
\end{figure}
Once the association information is known, we formulate an optimization problem to find Pose and Shape of the Object. Ceres \cite{agarwal2015others} toolbox is used for this purpose. In following subsections we take a look at different constraints used in the formulation.

The final cost function is 
\begin{equation} \label{eq:ceres_cost_total}
\Phi = \Phi_{pose} + \Phi_{normal} + \Phi_{shape}
\end{equation}

\subsubsection{pose constraints}
In figure \ref{Fig:ceres_line_resection}, $AB$ is a 3D line projected to an image plane forming $ab$ 2D line. The normal constructed by the $cross$ $product$ of $oa$ and $ob$ is perpendicular to the 3D line. Let $M$ be the point on the line and $D$ be the direction.  

\[ \hat{N} = \overline{oa} \otimes \overline{ob} \]
\begin{equation} \label{eq:normal_constraint2}
	\mathbf{\hat{N} \odot (R\times M + T) = 0}
\end{equation}

taking the difference between two points $M_1$ and $M_2$ from same line
\[ \hat{N} \odot(R\times(M_1 - M_2) = 0\]
\begin{equation} \label{eq:normal_constraint1}
	\mathbf{\hat{N} \odot{R\times D} = 0}
\end{equation}

So, the cost function is

\begin{equation} \label{eq:phi_pose}
	\mathbf{
    \Phi_{pose} = \sum_{l_i \in X} ||\hat{N} \odot{(R\times D_i)}||^2 + ||\hat{N} \odot{(R\times M_i + T)}||^2
    }
\end{equation}

R and T are the parameters we want to optimize for.

\subsubsection{normal constraint}
Each category object has a base e.g base of chair for sitting. We define base of the object as the plane which is parallel to ground plane when the object is kept in normal position.

We use this observation and put in the constraint to force the base of object to be parallel to ground. We consider normal of ground plane to be the y-axis.

\begin{equation} \label{eq:phi_normal}
	\mathbf{\Phi_{normal} = \hat{Y} - (R\times M_1 + T)\otimes(R\times M_2 + T)}
\end{equation}

here $\hat{Y}$ is the y-axis and $M_1$ and $M_2$ belongs to the adjacent base lines from $X$.

\subsubsection{shape constraints}
Finally, we use our eigen vector formulation discussed in section \ref{subsection:line_parameterization} to optimize for the shape of object. 
\[ X = \overline{X} + V\Lambda \]
So, for any line $ L_i \in X$
\[ L_i = \overline{L_i} +  \sum_{j in size(V)} \lambda_j \times L_{ji}\]
expanding $L_i$ = $<M_i, D_i>$
\begin{equation} \label{eq:shape_mid}
 \mathbf{M_i = \overline{M_i} + \sum_{j} \lambda_j \times M_{ji} }
\end{equation}
\begin{equation} \label{eq:shape_dir}
 \mathbf{D_i = \overline{D_i} + \sum_{j} \lambda_j \times D_{ji} }
\end{equation}

using these in equation \ref{eq:phi_pose} to get shape constraint 
\begin{equation} \label{eq:phi_shape}
\begin{split}
\Phi_{shape} = \sum_{l_i \in X}(||\hat{N}\odot{(R\times(\overline{D_i} + \sum_{j}\lambda_j\times D_{ji}))}||^2 \\  + ||\hat{N}\odot{(R\times (\overline{M_i} + \sum_{j}\lambda_j \times M_{ji}) + T)}||^2)
  \end{split}
\end{equation}

\subsubsection{Optimizing Pose and Shape} 
The optimizer is called for pose, $R$ and $T$, of the object with cost 
\begin{equation}
\label{eq:phi_pose}
\Phi = ||\Phi_{pose}||^2 + ||\Phi_{normal}||^2
\end{equation}
followed by the call to optimizer for shape, $\lambda's$, of the object with cost
\begin{equation}
\label{eq:phi_shape}
\Phi = ||\Phi_{shape}||^2 + \rho(\wedge)
\end{equation}
where $\rho(\wedge)$ is a regularizer that prevent shape parameters ($\wedge$) from deviating from the category-model.
Improvement in shape can result in improvement of the pose of object and vice-versa, thus, both optimizations are called iteratively to achieve better results.

\subsection{Integrating Object Pose with Monocular SLAM}
The category-models learned using \emph{line} based approach are incorporated into a monocular SLAM back end. Here we have, $\mathbf{Z}_{{ij}}$ = $\mathbf{Z}_{j}$$\mathbf{{Z}_{i}}^{-1}$, where $\mathbf{Z}_{{ij}}$  $\in$ SE(3) represents rigid-body transform of a 3D point in camera frame at time \emph{i} with respect to camera frame at time \emph{j}. $\mathbf{Z}_{{ij}}$ is a 4$\times$4 matrix represented below
\begin{equation}
\mathrm{Z_{ij}} = \begin{bmatrix}
     \mathrm{R_{ij}} & \mathrm{t_{ij}} \\
     0 & 1 
     \end{bmatrix}
     \quad \textrm{where} \; \mathrm{R_{ij}}\; \epsilon \; SO(3), \; t_{ij} \; \epsilon \; \mathbb{R}^{3}
\end{equation}
If 3D coordinate of a world point $^w$\emph{X} with respect to frame \emph{i} is $^i$\emph{X} then using the transformation $\mathbf{Z}_{{ij}}$ we can represent it with respect to camera frame \emph{j} as $^j$\emph{X} = $\mathbf{{Z}^{i}}_{{ij}}$
\emph{X}.

For a given set of relative pose measurement \{$\mathbf{\overline{Z}}_{{ij}}$\} of robot across all the frames $\forall$ \emph{i} \emph{j} $\in$ $\{1 \dots F\}$, we define the pose-SLAM problem as estimating $\mathbf{Z}_{i}$ $\forall$ \emph{i} $\in$ $\{1 \dots F\}$ that maximizes the log-likelihood of relative pose measurements, which can be framed as problem of minimizing observation errors (minimizing the negative of log likelihood).  

\begin{equation} \label{eq:pose_optimization}
\min_{Z_{i},i\epsilon\{1..F\}} \varepsilon_{pose} = \sum_{\overline{Z}_{ij}} {\| Log(\overline{Z}^{-1}_{ij} Z_{j} Z^{-1}_{i}) \|}_{\sum_{ij}} 
\end{equation}
Where $\sum_{ij}$ is assumed to be the uncertainty associated with each pose measurement $\mathbf{\overline{Z}}_{{ij}}$. 
In order to minimize the problem posed above(\ref{eq:pose_optimization}), we employ factor graphs \cite{kaess2012isam2} using publicly available GTSAM \cite{dellaert2012gtsam} framework to construct and optimize the proposed factor graph model.

Minimizing error function (\ref{eq:phi_pose}) and (\ref{eq:phi_shape}) in an alternating manner with respect to object shape and pose parameters yield estimated shape($\wedge$) and pose(${\overline{Z}_{i}}^{O}$) for a given frame \emph{i}. Pose observation obtained after shape and pose error minimization form additional factors in SLAM factor graph, therefore for each object node in the factor graph if  pose of object $O_{\phi(m)}$ is denoted by $Z^{O_{\phi(m)}}$, following error is minimized.
\begin{equation}
\min_{\underset{{\overline{Z}_{i}}^{O_{n}}, n \epsilon \{1..N\}}{Z_{i},i\epsilon \{1..F\}}} \varepsilon_{obj} = \sum_{i=1}^{F}\sum_{m=1}^{M} {\| Log(\left(\overline{Z}^{i}_{O_m}\right)^{-1} Z^{-1}_{i} Z^{O_{\phi(m)}} ) \|} 
\end{equation}
Here $\phi(m)$ denotes data association function that uniquely identifies every object $O_{m}$ observed so far.
Finally object-SLAM error $\varepsilon$ that jointly estimate robot pose and object poses using relative object pose observations is expressed as:
\begin{equation}
\varepsilon = \varepsilon_{pose} + \varepsilon{obj}
\end{equation}

\section{Results}
\begin{figure*}[!htb] 
 \centering
  \includegraphics[width=1\linewidth, height=2.75cm]{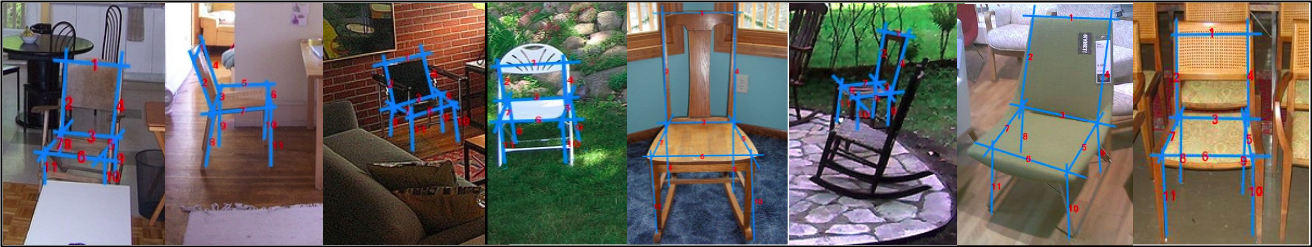}
  \caption{Results of our pipeline obtained on real chairs from PASCAL VOC \cite{everingham2010pascal} dataset.}
  \label{fig:pascal_chairs}
\end{figure*}

In this section we present experimental results on multiple real-world sequences comprising of different category objects vis-a-vis chair, table, and Laptop . We evaluate the performance of the proposed \emph{line} based approach for Object SLAM. We also emphasize on the nature of our approach that exploit Key edges in an object, corresponding to the respective wire-frame model to obtain object trajectory and precisely estimate their pose in various real-world scenarios. Fig \ref{fig:pascal_chairs} shows result of our \emph{Line} based pipeline on PASCAL VOC \cite{everingham2010pascal} dataset.

In Table \ref{tab:quantitative_result_table}, the comparison of our approach against the trajectory generated by ORB-SLAM is shown. The localization error is computed for each object and best, worst and average are reported. Our objects CAD models are in metric scale and we scale the trajectory using ration of translation between end points in trajectory. After doing this, the results generated are in meters. The ground truth is collected by placing markers at the object positions. This table is to emphasize that our approach is able to embed objects in 3D space without deteriorating (even slightly improving it) the trajectory generated by ORB SLAM.

Lastly, we evaluate our pipeline against the keypoint method\cite{parkhiya2018constructing} by comparing the execution times. The time bottleneck for keypoint method during evaluation is in the forward pass of network.

Here, we compare frame processing time for both method for a $856\times480$ image containing 3 objects. The hardware specifications for keypoint method are TitanX GPU with 12 GB memory and for line based method intel i5 processor with 8GB ram.

Time per frame in keypoint method \\
\phantom{hspace*ab} = $3 \times$ inference time per object\\
\phantom{hspace*ab} = $3 \times$ 285 ms\\
\phantom{hspace*ab} = $855 ms$\\
\phantom{hsp}Time per frame in our method \\
\phantom{hspace*ab} = time per frame for LSD + $3 \times$ processing time per image\\
\phantom{hspace*ab} = $0.25 + 3\times$ 120\\
\phantom{hspace*ab} = $360.25 ms$

So, we have a increase in speed by more that 2 times for the same process.

Futher, we provide a video run and other relevant results in the supplementary material.

\subsection{Dataset}
\begin{figure*} 
\centering
\begin{minipage}[t]{0.19\textwidth}
\includegraphics[width=\textwidth, height=5.5cm]{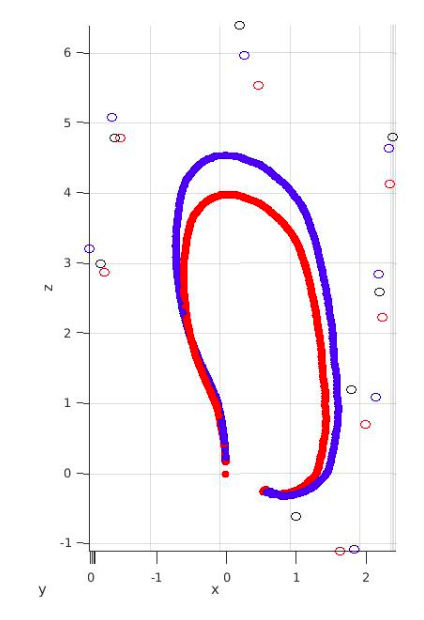}
\begin{center}
(a) Seq 1 without OLC
\end{center}
\end{minipage}
\begin{minipage}[t]{0.19\textwidth}
\includegraphics[width=\textwidth, height=5.5cm]{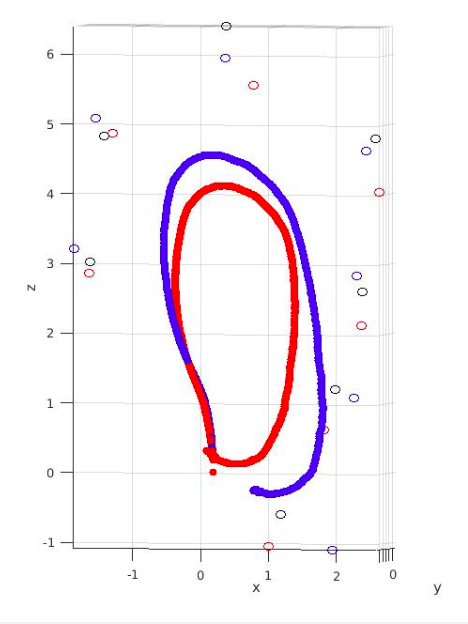}
\begin{center}
(b) Seq 1 with OLC
\end{center}
\end{minipage}
\begin{minipage}[t]{0.19\textwidth}
\includegraphics[width=\textwidth, height=5.5cm]{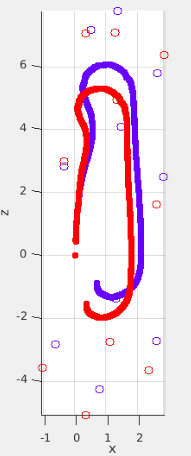}
\begin{center}
(c) Seq 2 without OLC
\end{center}
\end{minipage}
\begin{minipage}[t]{0.19\textwidth}
\includegraphics[width=\textwidth, height=5.5cm]{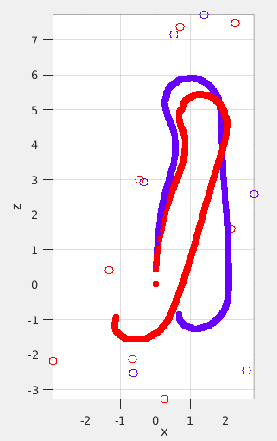}
\begin{center}
(d) Seq 2 with OLC
\end{center}
\end{minipage}
\begin{minipage}[t]{0.20\textwidth}
\includegraphics[width=\textwidth, height=5.5cm]{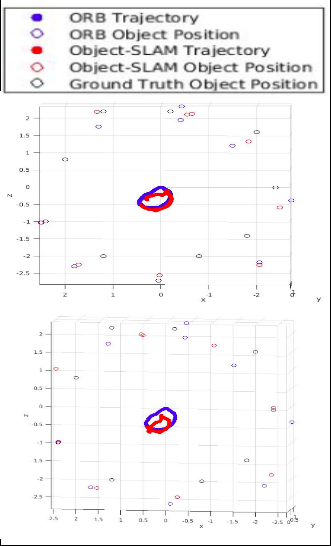}
\begin{center}
(e) Seq 3 without and with OLC
\end{center}
\end{minipage}

\caption{Shown here are estimated trajectories and object location for different monocular sequences using ORB-SLAM and Object-SLAM with and without Object Loop Closure(OLC). }
\label{Fig:sequence_trajectory}
\end{figure*}

We demonstrate object SLAM using our approach on numerous sequences of monocular video in an indoor setting, comprising of office spaces and laboratory which constitute our dataset. We collected our dataset using a micro aerial vehicle (MAV) flying at a constant height above the ground.
Sequence 1 and 2 of our dataset are elongated loops with 2 parallel sides, following dominant straight line motion while Sequence 3 is a 360$^{\circ}$ rotation in place with no translation from origin. Estimated robot(MAV) trajectory and object locations for these runs have been visualized in Fig \ref{Fig:sequence_trajectory} for both ORB-SLAM \cite{ORB} and our \emph{line} based object-SLAM, with and without object loop closure.  

\begin{table*}[]
\caption{Quantitative results on different sequences} 
\label{tab:quantitative_result_table}
\begin{tabular}{|l|l|l|l|l|l|}
\hline
\multirow{2}{*}{Sequence ID} & \multirow{2}{*}{Approach} & \multirow{2}{*}{\# Objects} & \multicolumn{3}{l|}{Object localization Error (metres)} \\ \cline{4-6} 
 &  &  & Best & Worst & Average \\ \hline
\multirow{2}{*}{1} & ORB & \multirow{2}{*}{7} & 0.1558 & 1.0331 & 0.457 \\ \cline{2-2} \cline{4-6} 
 & Ours &  & 0.1592 & 0.9190 & 0.5030 \\ \hline
\multirow{2}{*}{2} & ORB & \multirow{2}{*}{12} & 1.52 & 3.20 & 2.23 \\ \cline{2-2} \cline{4-6} 
 & Ours &  & 1.55 & 3.12 & 2.1 \\ \hline
 \multirow{2}{*}{3} & ORB & \multirow{2}{*}{9} & 3.05 & 4.65 & 3.89 \\ \cline{2-2} \cline{4-6} 
 & Ours &  & 3.75 & 4.61 & 3.85 \\ \hline
\end{tabular}
\end{table*}

\subsection{Instance Retrieval}
We apply principle component analysis \cite{jolliffe2011principal} to select Eigen vectors that represent the object space in section 3.2. In section 3.4, we formulate the optimization problem to solve for the shape of the object. The solution for this optimization gives us coefficients of the top Eigen vectors, which represent shape of the object.

\begin{figure*}[!htb] 
 \centering
  \includegraphics[width=1\linewidth, height=2.8cm]{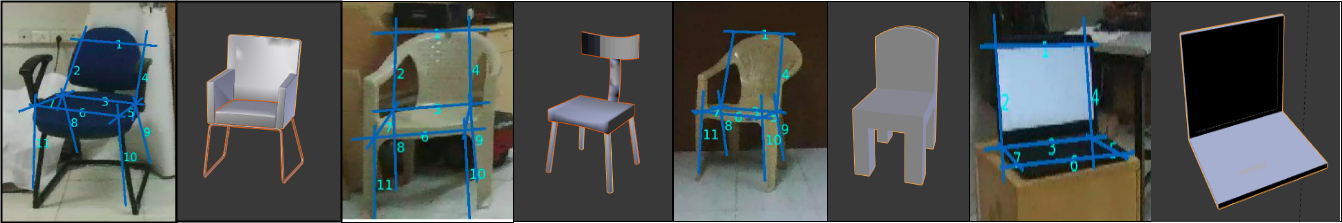}
  \caption{Instance retrieval results for objects. Query images and corresponding retrieved CAD models.}
  \label{fig:instance_retrieval}
\end{figure*}

Now we retrieve the closest instance from the 3D CAD model collection, that best defines the shape of the object, by running a K-Nearest Neighbors search. In Fig. \ref{fig:instance_retrieval} we present results of retrieving instance of 3D CAD model by running a 5-Nearest Neighbor search and then manually selecting the closest instance. We used these retrieved instances to visualize objects in a robot trajectory. 

\subsection{Normal Correction}
\begin{figure}[!htb] 
 \centering
  \includegraphics[width=1\linewidth, height=4cm]{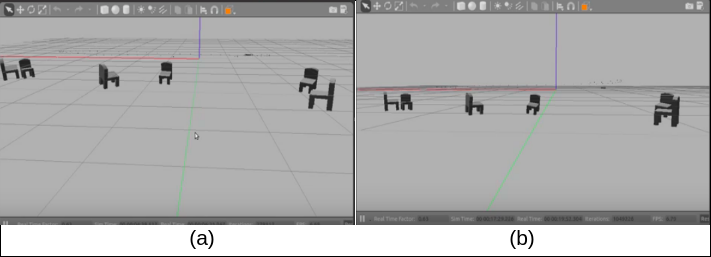}
  \caption{Image (a) shows trajectory visualisation without incorporating normal cost in our optimization function whereas (b) shows the difference when normal cost is included. }
  \label{Fig:normal_cost}
\end{figure}

In the pose optimization formulation, while solving for pose, \emph{R} and \emph{T} of an object a normal correction cost($\phi_{normal}$) is  also included. In fig \ref{Fig:normal_cost} clear improvement can be seen in pitch and roll of the objects with the inclusion of the normal cost(eq. \ref{eq:phi_normal}), herein demonstrated using trajectory corresponding to sequence 3 of our dataset visualized with gazebo.

\section{Conclusions}
The paper introduces a novel line based parameterization to represent various objects that are generally available in indoor environment. We provide a complete pipeline which finds object poses using Pose and Shape optimization and then embeds the objects in map with the monocular SLAM trajectory, using factor graph optimization backend, to localize the object with reasonable accuracy in the navigable space.

We show the result of the proposed pipeline on various real world scenes containing objects from multiple category. The pipeline is able to localize the objects in map without deteriorating the ORB SLAM performance and in fact improving the trajectory to some extent.

The line based parameterization can prove to be useful in cases where keypoint information is hard to obtain. It circumvents the training and data collection phases and speeds up the evaluation process for associaton.

The performance of pipeline depends on robustness of the association algorithm. We plan to implement a graph based optimization method to give the associations for objects and further improve the performance and robustness of the proposed pipeline.